\documentclass[sigconf,authorversion,nonacm]{acmart}
\usepackage{algorithm} 
\usepackage{algpseudocode}
\usepackage[algo2e]{algorithm2e} 
\usepackage{graphicx}
\usepackage{multicol,caption}
\usepackage{lipsum}

\AtBeginDocument{%
  \providecommand\BibTeX{{%
    \normalfont B\kern-0.5em{\scshape i\kern-0.25em b}\kern-0.8em\TeX}}}

\acmConference[AdvML '20]{AdvML '20: Workshop on Adversarial Learning Methods for Machine Learning and Data Mining}{August 24, 2020}{San Diego, CA}
\acmBooktitle{AdvML '20: Workshop on Adversarial Learning Methods for Machine Learning and Data Mining, August 24, 2020, San Diego, CA}

\begin{document}

\title{Identifying Audio Adversarial Examples via Anomalous Pattern Detection}

\author{Victor Akinwande}
\affiliation{\institution{IBM Research Africa}}
\authornote{Correspondence: victor.akinwande1@ibm.com}

\author{Celia Cintas}
\affiliation{\institution{IBM Research Africa}}

\author{Skyler Speakman}
\affiliation{\institution{IBM Research Africa}}

\author{Srihari Sridharan}
\affiliation{\institution{IBM Research Africa}}






\begin{abstract}
Audio processing models based on deep neural networks are susceptible to adversarial attacks even when the adversarial audio waveform is 99.9\% similar to a benign sample.
Given the wide range of applications of DNN-based audio recognition systems from automotives to virtual assistants, detecting the presence of adversarial examples is of high practical relevance.
We propose a method to detect audio adversarial samples. Employing anomalous pattern detection techniques in the activation space of these models, we show that 2 of the recent and current state-of-the-art adversarial attacks on audio processing systems systematically lead to higher-than-expected activation at some subset of nodes and we can detect these with up to an AUC of $0.98$ with no degradation in performance on benign samples. Furthermore, our work strengthens the study of properties of adversarial examples that hold across multiple domains.
\end{abstract}



\maketitle

\section{Introduction}
\label{sec:intro}

Speech-based interaction is widely used in virtual personal assistants (e.g. Siri, Google Assistant) and also moving to more critical areas such as virtual assistants for physicians (e.g. Saykara). Given the increasing application of deep neural network-based audio processing systems, the robustness of these systems is of high relevance. Neural networks are susceptible to adversarial attacks where an adversarial example \citep{szegedy2013intriguing} typically crafted by adding small perturbations to the inputs causes an erroneous output that may be prespecified by the adversary \citep{carlini2017towards}. Existing work on adversarial examples focuses predominantly on image processing tasks \citep{akhtar2018threat} and recently more attention is being given to other domains such as audio for automatic speech recognition (ASR) \citep{alzantot2018did, gong2017crafting, carlini2018audio, taori2018targeted, gong2017crafting, qin2019imperceptible}. Thus, the detection of adversarial attacks is a key component in building robust models. 

Given an input audio waveform $x$, an ASR system $f(.)$, and a target transcription $y$, most attacks seek to find a small perturbation $\delta$ such that $x' = x + \delta$ and $f(x') = y$ though $f(x) \neq y$. This is referred to as a targeted attack and such an adversarial audio waveform may be 99.9\% similar to a benign sample \citep{carlini2018audio}. Also, recent work \citep{schonherr2019robust, qin2019imperceptible, yakura2018robust} has demonstrated the feasibility of these adversarial samples being played over-the-air by simulating room impulse responses and making them robust to reverberations.
We observe that the key differentiation between generating adversarial examples across different tasks or input modalities such as images, audio or text lies in a change of architecture as these attacks generally attempt to maximize the training loss and it is valuable to study properties of adversarial examples that hold across multiple domains. Whereas existing work in defending against audio adversarial attacks in propose preprocessing the audio waveform as a means of defense \citep{yang2018characterizing, das2018adagio, rajaratnam2018isolated, subramanian2019robustness}, our work treats the problem as anomalous pattern detection and operates in an unsupervised manner without apriori knowledge of the attack or labeled examples. We also do not rely on training data augmentation or specialized training techniques and can be complementary to existing preprocessing techniques.

We claim two novel contributions of this work. First, we propose a detection mechanism by using nonparametric scan statistics, efficiently optimized over node-activations to quantify the anomalousness of such inputs into a real-valued ``score''.
Second, we show empirical results across two state-of-the-art audio adversarial attacks ~\citep{carlini2018audio, qin2019imperceptible} with consistent performance comparable to the current state-of-the-art without losing accuracy of benign samples which is a known downside of preprocessing approaches.

\section{Related Work}
The property of deep neural networks being susceptible to adversarial attack was largely established in \citep{biggio2013evasion} and \citep{szegedy2013intriguing}. Since then, numerous kinds of attacks have been designed across multiple data modalities including images \citep{goodfellow2014explaining, papernot2016limitations, carlini2017towards} and audio \citep{cisse2017houdini, gong2017crafting, alzantot2018did, carlini2018audio, qin2019imperceptible}. A common kind of attack is an evasion attack \citep{chakraborty2018adversarial} where the adversary with no influence over the training of the model attempts to evade the system during inference by crafting malicous samples. Early work on audio adversarial research focused on untargeted attacks where the goal is to produce an incorrect but arbitrary transcription for an ASR given an input that has been minimally perturbed. In \citep{cisse2017houdini}, a multi-modal attack is introduced that proposes the use of a differentiable surrogate loss function in place of the task loss, and applied to speech recognition (with phonetic constraints). However, targeted attacks were notably difficult. In \citep{carlini2018audio}, an iterative optimization targeted attack is introduced with 100\% success rate on state-of-the-art audio models with the limitation being the inability of these samples to remain adversarial when played over-the-air. In \citep{qin2019imperceptible}, this limitation is addressed by leveraging psychoacoustics towards more imperceptible and over-the-air attacks.

On the other hand, defending against audio adversarial attacks has predominantly focused on preprocessing techniques such as mp3 compression, quantization, adding noise, or smoothing \citep{rajaratnam2018isolated, das2018adagio, yang2018characterizing, subramanian2019robustness}. However, these approaches modify the input in some way and affect performance on benign samples. Particularly,~\citep{yang2018characterizing} propose using the explicit temporal dependency in audio e.g. correlations in consecutive waveform segments and show that it is affected by adversarial perturbations. However, there exists an inductive bias in form of what time step(s) to break up the waveform sequence for it to be evaluated as being adversarial while minimizing the performance degradation if the sample is benign. Also, for an audio sample to be deleterious, in the real world, only a small subset of the transcription given a waveform sequence needs to be adversarially targeted. For example, a sample that translates to - ``Alexa, please call the doctor'' and changed to ``Alexa, please call the doorman'' using a combination attack may have a low Word Error Rate in the face of Temporal Dependency. A different approach to detecting adversarial samples is to leverage the uncertainty estimates of a trained model under different dropout configurations \citep{feinman2017detecting}. In \citep{jayashankar2020detecting}, they extend this defense to an ASR model with strong performance but only evaluate with a simple model whereas the current state-of-the-art attack is applied to more complex sequence-sequence based models where naive dropout is known to significantly affect performance \citep{bayer2013fast}.

This motivates the need to explore other mechanisms for adversarial audio detection. Our work shows strong discriminative power against adversarial samples without preprocessing the input thus preventing performance degradation on clean samples and we apply our defense to a sequence-sequence based model with attention, Lingvo ASR system \citep{shen2019lingvo} with competitive results.

\section{Non-parametric Scan Statisitcs and Subset Scanning }
\label{sec:method}

Subset scanning (SS) is an approach to pattern detection, which treats the problem as a search for the ``most anomalous'' subset of observations in the data. Herein, anomalousness is quantified by a scoring function, $F(S)$. We formulate the audio adversarial detection problem as being able to efficiently identify $S^{\ast} =\arg\max_{S}F(S)$ over all relevant subsets of node activations within an ASR that is processing audio waveforms at runtime. We use non-parametric scan statistics (NPSS) that have been used in other pattern detection methods ~\citep{neill-npss-2007, mcfowland-fgss-2013, mcfowland-tess-2018, feng-npss_graph-2014} as our scoring function.

Let there be $M$ clean audio samples $X_z$ included in $D_{H_0}$. These samples generate activations $A^{H_0}_{zj}$ at each node $O_j$. Let $X_i$ (not in $D_{H_0}$) be a test sample under evaluation. This audio sample creates activations $A_{ij}$ at each node $O_j$ in the network. The $p$-value, $p_{ij}$, is the proportion of background activations $A^{H_0}_{zj}$ greater than the activation induced by the test sample $A_{ij}$ at node $O_j$. We convert the test sample $X_i$ to a vector of $p$-values $p_{ij}$ of length $J = |O|$.

The key assumption is that under the alternative hypothesis of an anomaly present in the activation data, then at least some subset of the activations $S_O \subseteq O$ will systematically appear extreme. We now turn to non-parametric scan statistics to identify and quantify this set of $p$-values. 

The general form of the NPSS score function is 
\begin{equation}
F(S)=\max_{\alpha}F_{\alpha}(S)=\max_{\alpha}\phi(\alpha,N_{\alpha}(S),N(S)),
\end{equation}
where $N(S)$ represents the number of empirical $p$-values contained in subset $S$ and $N_{\alpha}(S)$ is the number of $p$-values less than (significance level) $\alpha$ contained in subset $S$. There are well-known NPSS that can be utilized \citep{mcfowland-tess-2018} such as the Kolmogorov-Smirnov test ~\citep{kolmogorov1933sulla} or Higher-Criticism \citep{donoho-gof-2004}. In this work, we use the Berk-Jones test statistic \citep{berk-bj-1979} $\phi_{BJ}$.

\begin{equation}
\phi_{BJ}(\alpha,N_\alpha,N) = N*{KL} \left(\frac{N_\alpha}{N},\alpha\right), 
\end{equation}
where $KL$ is the Kullback-Liebler divergence $KL(x,y) = x \log \frac{x}{y} + (1-x) \log \frac{1-x}{1-y}$ between the observed and expected proportions of significant $p$-values. Berk-Jones is a log-likelihood ratio for testing whether the $p$-values are uniformly distributed on $[0,1]$ as compared to a piece-wise constant alternative distribution, and has been shown to fulfill several optimality properties.

\subsection{Efficient Maximization of NPSS}
Although NPSS provides a means to evaluate the anomalousness of a subset of node activations $S_O$ discovering which of the $2^J$ possible subsets provides the most evidence of an anomalous pattern is computationally infeasible for moderately sized data sets. However, NPSS has been shown to satisfy the linear-time subset scanning (LTSS) property \citep{neill-ltss-2012}, which allows for an efficient and exact maximization over subsets of data. 

The LTSS property uses a priority function $G(O_j)$ to rank nodes and then proves that the highest-scoring subset consists of the ``top-k'' priority nodes for some $k$ in $1 \ldots J$. The priority of a node for NPSS is the proportion of $p$-values that are less than $\alpha$. However, because we are scoring a single audio sample and there is only one $p$-value at each node, the priority of a node is either 1 (when the $p$-value is less than $\alpha$) or 0 (otherwise).  Therefore, for a fixed, given $\alpha$ threshold, the most anomalous subset is all and only nodes with $p$-values less than $\alpha$. We refer to this $\alpha$ threshold as $\alpha_{max}$.

To maximize the scoring function over $\alpha_{max}$, let $S_{(k)}$ be the subset containing the $k$ nodes with the smallest $p$-values and $\alpha_k$ be the largest $p$-value among these $k$ nodes. The LTSS property guarantees that the highest-scoring subset (for the chosen $\alpha_{max}$) will be one of these $J$ subsets ${S_{(1)}, S_{(2)}, \ldots S_{(J)}}$ with their corresponding $\alpha_k$ threshold. Any subset of nodes that does not take this form (or uses an alternate $\alpha_k$) is provably sub-optimal and not considered. Critically, this drastically reduced search space still guarantees identifying the highest-scoring subset of nodes for a test audio sample under evaluation. Pseudo-code for subset scanning over activations for audio samples can be found in Algorithm~\ref{alg:algo_disjdecomp}.

\algrenewcommand\algorithmicrequire{\textbf{Input:}}
\algrenewcommand\algorithmicensure{\textbf{Output:}}

\begin{algorithm}[t]
\begin{algorithmic}

\SetKwFunction{TrainNetwork}{TrainNetwork}
\SetKwFunction{ExtractActivation}{ExtractActivation}
\SetKwFunction{SortAscending}{SortAscending}
\SetKwFunction{NPSS}{NPSS}

\Require Background samples: $X_z \in D^{H_0}$, Evaluation sample: $X_i$, $\alpha_{\text{max}}$. $M = |D^{H_0}|$.
\BlankLine\\
$ ASR \leftarrow$ \TrainNetwork(training dataset)\\
$ ASR_y \leftarrow$ Some flattened layer of $ASR$
\BlankLine
\For{$z\leftarrow 0$ \KwTo $M$}{
\For{$j\leftarrow 0$ \KwTo $J$}{
 $A^{H_0}_{zj} \leftarrow $ \ExtractActivation($ASR_{y}$, $X_z$)
}
}
\For {$j\leftarrow 0$ \KwTo $J$}{
$A_{ij} \leftarrow $ \ExtractActivation($ASR_{y}$, $X_i$)
}
$p_{ij} = \frac{\sum_{X_z \in D^{H_0}} I(A_{zj} >= A_{ij} ) + 1}{M+1}$

$p^{\ast}_{ij} = \{y < \alpha_{\text{max}} \: \forall \: y \subseteq p_{ij} \}$

$p^s_{ij} \leftarrow $ \SortAscending($p_{ij}$)

\For {$k\leftarrow 1$ \KwTo $J$}{
$S_{(k)} = \{p_y \subseteq p^s_{ij} \forall y \in \{1, \ldots, k\} \}$

$\alpha_{k} = max(S_{(k)})$

$F(S_{(k)}) \leftarrow $ \NPSS($\alpha_{k}$, k, k)\; 
}

$k^{\ast} \leftarrow \arg\max F(S_{(k)})$

$\alpha^{\ast} = \alpha_{k^\ast}$\;

$S^\ast = S_{(k^\ast)}$\;

\Return $S^\ast$, $\alpha^{\ast}$, and $F(S^\ast)$
 \end{algorithmic}
 
\caption{Pseudo-code for subset scanning over activations of individual audio samples. For a given audio sample, we compute an anomalousnes score $F(S^\ast)$ that characterizes how extreme the node-activations for that sample is compared to a background distribution. We also return the subset of nodes $S^\ast$ and alpha $\alpha^{\ast}$ that maximize the score.}
\label{alg:algo_disjdecomp}

\end{algorithm}

\section{Experiments}
\label{sec:expres}

In this section, we introduce the datasets, target models, and attack types evaluated in our proposed approach. We describe the experimental setup for three experiments: DeepSpeech model evaluated with Common Voice dataset, DeepSpeech evaluated with Librispeech dataset and Lingvo evaluated with Librispeech. We summarize our results in Table \ref{tab:auc} and compare with the current state-of-the-art detection methods.

\subsection{Datasets}

\textbf{Mozilla Common Voice dataset:} Common Voice is an audio dataset provided by Mozilla. This dataset is public and contains samples from voice recordings of humans. We resample the subset used in our experiments to 16Khz with an average duration of 3.9 seconds.\\
\textbf{Dataset LibriSpeech dataset:} LibriSpeech \citep{panayotov2015librispeech} is a corpus of approximately 1000 hours of 16Khz English speech derived from audiobooks from the LibriVox project. Samples in our subset have an average duration of 4.3 seconds.

\subsection{Target Models and Attacks} 

\textbf{DeepSpeech:} We apply CW attack \citep{carlini2018audio} on version 0.4.1 of DeepSpeech \citep{hannun2014deep}.
To generate adversarial examples on ASR systems ~\citep{carlini2018audio} use the typical adversarial example generation algorithm that solves Eq.~\ref{eq:carnili}. Particularly, they set $\ell$ to the CTC-loss and use the max-norm which has the effect of adding a small amount of adversarial perturbation consistently throughout the audio sample.

\begin{equation}
\begin{aligned}
&\text { min } \ell(f(x+\delta), y) +\alpha \cdot\|\delta\|\\
&\text { such that }\|\delta\|<\epsilon
\end{aligned}
\label{eq:carnili}
\end{equation}

We set the number of iterations to 100 with a learning rate of 100 and generate adversarial examples for Mozilla Common Voice (the first 100 test instances) and Librispeech (the first 200 test-clean instances) with a success rate of 92\% and 94.5\% respectively.\\
\textbf{Lingvo:} \citep{qin2019imperceptible} proposed a new attack (IA) improving the CW attack leveraging frequency masking by enforcing the power spectral density $p_\delta$ of the perturbation in the short-time Fourier transform (STFT) domain to be under a frequency masking threshold $\theta_x$ of the original audio sample by optimizing the following:

\begin{equation}
\begin{aligned}
    &\text { min } \ell(f(x+\delta), y) + \alpha \sum_{k=0}^{\lfloor \frac{N}{2}\rfloor} \max\{p_\delta(k) - \theta_x(k), 0\},
\end{aligned}
\label{eq:qin}
\end{equation}
where %
$\alpha$ controls the relative importance of the term making the perturbation imperceptible, and $N$ is the STFT window size. The attack is a 2 stage process whereby $\alpha=0$ during the optimization to find a perturbed sample transcribing as target $y$, and $\alpha$ is slowly increased to satisfy the imperceptibility constraint by fine-tuning the perturbation in the second stage. We apply IA attack on the Lingvo system \citep{shen2019lingvo} with a stage 1 learning rate of 100 with 1000 iterations and a stage 2 learning rate of 0.1 with 4000 iterations. We generate adversarial examples for Librispeech (the first 130 test-clean instances) with a 100\% success rate. 

\subsection{Experimental setup }

For each of our experiments, we extract the node activations, post activation function. In this case, after the relu function (NPSS can be applied on any activation function and architecture). We then extract activations from clean audio samples that form our distribution of activations under a null hypothesis of no adversarial noise present. Activations from samples in the evaluation set are compared against the activations from the clean samples to create empirical $p$-values for each sample. These $p$-values are scored by non-parametric scan statistics to quantify the anomalousness of each sample in the evaluation set (See Section \ref{sec:method}). In other words, each audio sample is quantified with an anomalousness score. We set $\alpha_{max}$ to $0.5$ for all experiments. Future supervised experiments could tune $\alpha_{max}$ to increase detection further. Once we obtain an anomalousness score for each sample, we evaluate the ability to separate clean samples from adversarial samples using AUC which is a threshold independent metric.\\
\textbf{DeepSpeech experiment on Common Voice:} We draw the first 1000 samples from the trainset and choose the first 800 as our background and remaining 200 as our clean samples. We randomly draw 90 samples from the 92 adversarially generated examples as our adversarial samples.\\
\textbf{DeepSpeech experiment on Librispeech:}, We draw the first 1000 samples from the test-clean set and choose the first 800 as our background and remaining 200 as our clean samples. We randomly draw 90 samples from the 189 adversarially generated examples as our adversarial samples.\\
\textbf{Lingvo experiment on Librispeech:} We draw the 201-600 samples from the test-clean set and choose the first 300 as our background and remaining 100 as our clean samples. We randomly draw 100 samples from the 130 adversarially generated examples as our adversarial samples. \\
As the audio samples vary in length the outputs from the activation nodes also vary, we choose the minimum across all sets (background, clean and adversarial) and cut off the activations at this length. Since both evaluated attacks perturb the entire audio waveform we believe the adversarial pattern will still be detected. We experiment with choosing different segments of the activation with no significant variance to AUC. \\
We compare against the ASR results of Temporal dependency (TD) \citep{yang2018characterizing}, and Dropout Uncertainty (DU) \citep{jayashankar2020detecting} which is only applied to DeepSpeech and evaluated using accuracy so we have no knowledge of how false positives impact it's performance.

\section{Results}

We show the results of scanning over specific layers of both evaluated models as well as the number of node activation for each layer in Table \ref{tab:auc}. Subset scanning (SS) achieves AUC as high as $0.973$ on Common Voice and $0.982$ on LibriSpeech with DeepSpeech (both in the "relu\_2" layer) and $0.755$ on LibriSpeech with Lingvo\\ ("fprop\_enc\_convl0\_relu0" layer). We leave the exploration of what layers provide the most discriminative potential for future work. These results show that subset scanning is indeed an effective method for detecting adversarial audio attacks. Given these results, we think that studying properties of adversarial examples that hold across multiple domains presents an interesting research direction.

\begin{table}[]
\caption{ AUC across attacks and datasets. We show the background (Bg), clean (Cl), and adversarial (Ad) sample sizes for each Model (M), Data (D) and Attack (A) triplet. We compare with current state-of-the-art detection methods Temporal Dependency (TD) and Dropout Uncertainty (DU) which uses accuracy (ACC) and show competitive detection across LibriSpeech (LS) and common voice (CV) for the CW and IA attacks on DeepSpeech (DS) and Lingvo (LV) models.}

\label{tab:auc}
\begin{tabular}{@{}p{2cm}p{2cm}p{1cm}p{1cm}p{1cm}@{}}
\toprule
{\raggedright M, D, A, \\(Bg, Cl, Ad)}& {\raggedright Layers \\dimensions} & {\raggedright TD \\(AUC)} & {\raggedright DU \\(ACC)} & {\raggedright SS \\(AUC)} \\ \midrule
DS, CV, CW      & 80, 2048 & 0.936 & 91.5 & 0.283 \\
(800, 200, 90)  & 80, 2048 &       & & 0.158 \\
                & 80, 4096 &       & & \textbf{0.973} \\
                & 80, 2048 &       & & 0.903 \\
DS, LS, CW      & 64, 2048 & 0.930 & NA & 0.568 \\
(800, 200, 90)  & 64, 2048 &       & & 0.038 \\
                & 64, 4096 &       & & \textbf{0.982} \\
                & 64, 2048 &       & & 0.527 \\
LV, LS, IA       & 179, 40, 32 & NA & NA & \textbf{0.755} \\
(300, 100, 100)  & 212, 20, 32 &  & & 0.491 \\
                 & 423, 40, 32 &  & & 0.571 \\
                 & 212, 20, 32 &  & & 0.479  \\ \bottomrule
\end{tabular}
\end{table}
\vspace{-0.05cm}
\section{Discussion}
An adversary with knowledge of the scoring function and $\alpha_{max}$ may be able to craft an attack with a loss function that optimizes an objective such that the node-activation $p$-values are uniformly distributed when compared to a background distribution while minimizing the perturbation to the sample. We briefly discuss the complexities of being able to achieve this. The attack should be marginally better than the CW attack as such, assuming the vanilla CW attack is enhanced with the addition of terms to defeat subset scanning the following would be required. First it should optimize across all the layers $L$ of the model, and secondly it should account for all $J$ subsets that the LTSS property guarantees the highest-scoring subset will be present in. The attack could also attempt to use a surrogate loss function as was proposed in \citep{cisse2017houdini} for non-differentiable and combinatorial loss functions, but we note that successful targeted attack was not achieved. Nonetheless, a simple but naive optimization can be formulated as in equation (\ref{eq:carnili-adaptive}).

Assuming this converges, it will take an enormous amount of time particularly if imperceptibility is desired (the IA attack takes 10hrs for a single sample on average on a machine with two Tesla K80 GPUs and 512Gb of RAM). In addition, recall that we set $\alpha_{max}=0.5$. We could also randomly choose $\alpha_{max}$ within a range that guarantees high detection on a class of attacks and further force an adaptive adversary to consider all possible values of $\alpha_{max}$.

\begin{equation}
\begin{aligned}
&\text { min } \ell(f(x+\delta), y) +\alpha \cdot\|\delta\| + \sum_{l=0}^{|L|}(\sum_{j=0}^{|J|}(SS(x) - SS((x+\delta)))\\
&\text { such that }\|\delta\|<\epsilon
\end{aligned}
\label{eq:carnili-adaptive}
\end{equation}

\section{Conclusion}
\label{sec:conclusion}
In this work, we proposed an unsupervised method for adversarial audio attack detection with subset scanning. Our method can detect multiple state-of-the-art adversarial attacks across multiple datasets. This discriminative potential comes from the idea that adversarially noised samples produce anomalous activations in neural networks that are detectable by efficiently searching over subsets of these activations. Whereas existing work in defending against audio adversarial attack proposes preprocessing the audio waveform as a means of defense, we treat the problem as anomalous pattern detection without apriori knowledge of the attack or labeled examples. We also do not rely on training data augmentation or specialized training techniques and can be complementary to existing preprocessing techniques. We note that an adversary may be able to use the knowledge of the technique to create an adaptive attack and discuss the challenges to achieving this. Future work will focus on leveraging the information contained in \emph{which subset} of nodes optimized the scoring function for that sample. This could lead towards new methods of neural network visualizations and explainability.

\clearpage

\bibliographystyle{ACM-Reference-Format}
\bibliography{main}

\end{document}